\documentclass[letterpaper, 10 pt, conference]{ieeeconf}  
\IEEEoverridecommandlockouts                              
\usepackage{graphicx} 
\usepackage{amsmath} 
\usepackage{amssymb}  
\usepackage[colorlinks=false]{hyperref}

\usepackage{caption}
\usepackage{subcaption}

\title{\LARGE \bf Learning Multi-Object Symbols for Manipulation with Attentive Deep Effect Predictors}

\author{Alper Ahmetoglu$^{1}$, Erhan Oztop$^{2,3}$, and Emre Ugur$^{1}$
\thanks{$^{1}$Department of Computer Engineering, Bogazici University
        {\tt\small \{alper.ahmetoglu\},\{emre.ugur\}@boun.edu.tr}}%
\thanks{$^{2}$Department of Computer Science, Ozyegin University
        {\tt\small erhan.oztop@ozyegin.edu.tr}}%
\thanks{$^{3}$OTRI, SISReC, Osaka University}
}

\begin{document}

\maketitle
\pagestyle{empty}

\begin{abstract}

In this paper, we propose a concept learning architecture that enables a robot to build symbols through self-exploration by interacting with a varying number of objects. Our aim is to allow a robot to learn concepts without constraints, such as a fixed number of interacted objects or pre-defined symbolic structures. As such, the sought architecture should be able to build symbols for objects such as single objects that can be grasped, object stacks that cannot be grasped together, or other composite dynamic structures. Towards this end, we propose a novel architecture, a self-attentive predictive encoder-decoder network with binary activation layers. We show the validity of the proposed network through a robotic manipulation setup involving a varying number of rigid objects. The continuous sensorimotor experience of the robot is used by the proposed network to form effect predictors and symbolic structures that describe the interaction of the robot in a discrete way. We showed that the robot acquired reasoning capabilities to encode interaction dynamics of a varying number of objects in different configurations using the discovered symbols. For example, the robot could reason that (possible multiple numbers of) objects on top of another object would move together if the object below is moved by the robot. We also showed that the discovered symbols can be used for planning to reach goals by training a higher-level neural network that makes pure symbolic reasoning.

\end{abstract}

\section{INTRODUCTION}
If a task of a robotic manipulator can be described in symbolic form, then the problem of obtaining a plan to achieve a task goal can be solved by reverting to search techniques from classical AI \cite{russell2010artificial}. In well-defined environments, a symbolic system describing the interactions of a robot with its environment can be designed manually. However, for unknown, unstructured and/or changing environments, it is desirable that the robot itself discovers the symbolic structures that are useful in reasoning and planning. As such, there is considerable literature on how the continuous sensorimotor experience of a robotic system can be converted to a set of symbolic interaction experiences \cite{Taniguchi2019}. Although the state-of-the-art in this frontier is progressing rapidly, currently, there is no learning system that can form symbols for a varying number of objects in one learning session. In our previous work \cite{ahmetoglu2020deepsym}, we have shown that the symbols formed with single object interactions can be used to bootstrap new symbol or rule formation while interacting with two objects. However, the transition from a single object to two objects required the construction of a new neural network. Furthermore, it was not possible to learn symbols from interactions that involve a varying number of objects, some of which may affect the action execution and others not. 

In this study, we aim to remove these limitations by acquiring a single neural system in order to discover symbols based on the sensorimotor data generated by the robot when it interacts with multiple objects with arbitrary multiplicity. To this end, we propose a deep neural architecture that includes self-attentive layers \cite{vaswani2017attention} with binary latent representations. We show the validity of the proposed architecture in a simulated manipulation scenario where a robotic arm interacts with object(s) while building symbolic representations. Most importantly, our system not only learns object-specific symbols but also learns multi-object symbols that are formed on-the-fly by automatically processing the related object symbols through the self-attention mechanism.
After learning, we investigated the formed symbols and observed that they are effective in making effect predictions. We showed that the learned symbols enable reasoning capabilities with multiple objects that may influence the interaction dynamics in various ways. We finally showed that the discovered symbols can be used for planning to reach desired goals, such as building desired structures with multiple objects via pure symbolic reasoning.

\begin{figure*}[t]
    \centering
    \includegraphics[width=\textwidth]{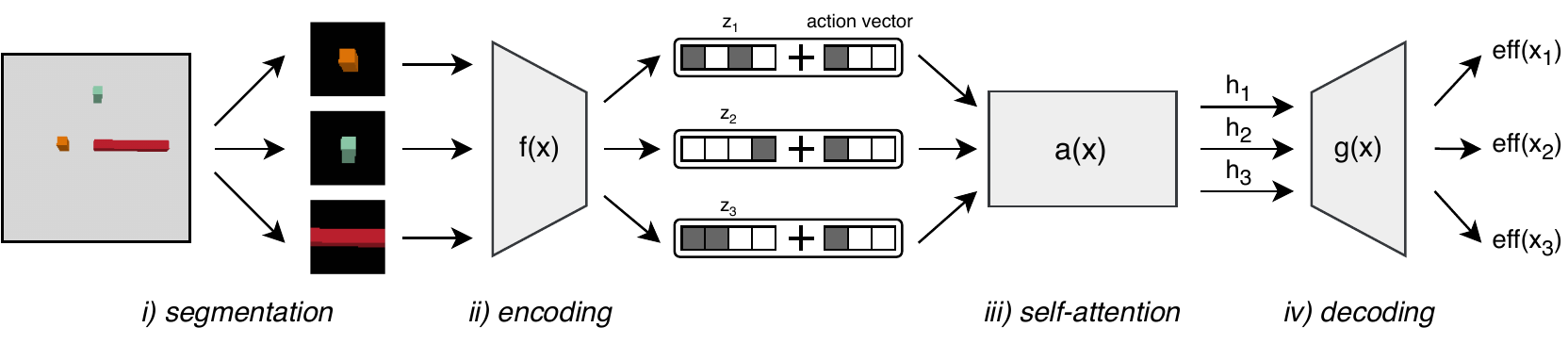}
    \caption{Attentive DeepSym architecture.}
    \label{fig:model}
\end{figure*}

\section{RELATED WORK}
Early symbol grounding studies in robotics (e.g., \cite{mourao2008,worgotter2009}) assumed the existence of manually defined symbols that were effective in plan generation. These studies collected data from interactions of agents and robots, and learned sensor to symbol mappings in order to ground the pre-defined symbols in the sensorimotor experience of the robot. In these studies, transition rules, which are connected by symbolic preconditions and effects were defined, and the continuous experience of the robot were used to map the manually defined symbolic predicates to the continuous perceptual space of the robots. Recently, \cite{dehban2022} proposed a deep neural network architecture that was based on Convolutional Variational Auto-Encoders to discover visual features that well-suit for pre-defined recognition and interaction tasks. \cite{lay2022unsupervised} used Multi-modal Latent Dirichlet Allocation (MLDA)  to learn the mapping between multi-modal sensory experience and preconditions and post-conditions of actions of a robot. We argue that pre-defining symbols in unknown and changing environments is not possible, and as stated by \cite{sun2000symbol} symbols should rather ``be formed in relation to the experience of agents, through their perceptual/motor apparatuses, in their world and linked to their goals and actions''.

Unsupervised discovery of discrete symbols and rule learning from the continuous sensorimotor experience of embodied agents has been recently studied in robotics in order to equip robots with advanced reasoning and planning capabilities \cite{Taniguchi2019,Konidaris2019}. \cite{ahmetoglu2022high} investigated the discovery of sub-symbolic neural activations that facilitate resource economy and fast learning in skill transfer but did not address high-level reasoning with discrete symbols. \cite{Konidaris2014,Konidaris2015} discovered symbols that were directly used as predicates in precondition and post-condition fields of action descriptors, which were represented in Problem Domain Definition Language (PDDL). This encoding allowed for making deterministic and probabilistic plans in 2-dimensional agent environments. The same architecture was extended to a real-world robotic environment in \cite{konidaris2018skills}, where symbols that represent absolute global states were learned and used for planning. \cite{james2019learning}, on the other hand, learned egocentric symbolic representations that enabled the agents to transfer the previously learned symbols to novel environments directly. Effect clustering techniques and SVM classifiers were used to discretize the continuous sensorimotor experience of the agents in these works. Whereas the previous work addressed learning symbols from given skills, \cite{silver2022learning} learned a set of skills from a set of symbolic predicates and a collection of demonstrations. \cite{Ugur-2015-ICRA,Ugur-2015-Humanoids} discovered discrete symbols and used these symbols in order to generate PDDL rules for planning by again combining effect clustering techniques to find discrete effect categories and SVM classifiers to discretize continuous object feature space. These studies used ad-hoc combinations of several machine learning methods. On the other, \cite{ahmetoglu2020deepsym} provided a more generic symbol formation engine, which used a novel deep network architecture that runs at the pixel level, and relied on purely predictive mechanisms in forming symbols instead of unsupervised clustering techniques. They used an effect predictor encoder-decoder network that took the object image and action as input, and exploited a binary bottleneck layer to automatically form object categories. Similar to this work, \cite{asai2017classical,asai2020learning,asai2021} also exploited deep neural networks with binary bottleneck units to find discrete state and effect symbols and achieve plan generation using these symbols. They realized symbolic planning in purely visual environments such as 2D puzzles, and they did not address actions and interactions of embodied agents.
While the majority of the previous work formed symbols to auto-encode the continuous state or make effective one-step-ahead effect predictions, \cite{silver2022inventing} proposed an architecture that learns state and action symbols that were explicitly optimized for effective and efficient multi-step planning. Different from our work, they assumed already defined goal predicates and a collection of demonstrations that enabled the robot to achieve the given goals.

Our ultimate aim with this model is to learn symbolic representations for objects by predicting the generated effect similar to the DeepSym architecture \cite{ahmetoglu2020deepsym}. The main difference from DeepSym is that the proposed architecture can model the interaction between a varying number of objects by allowing symbols to interact with each other via self-attention layers \cite{vaswani2017attention}. This not only allows object symbols to gather information from each other but also allows the architecture to be applicable to any number of interactions. For example, while the effect (in our case, the displacement of the object) of a push action to an object would require the symbol of the object (together with the action vector), the effect of stacking an object on top of another one would automatically require symbols of multiple objects. In the DeepSym architecture, this was only possible by manually concatenating cropped images of a fixed number of objects, which is not a scalable and general approach.

\section{METHOD}
\label{sec:method}
\begin{figure*}[tbp]
    \centering
    \begin{subfigure}[b]{0.3\textwidth}
        \centering
        \includegraphics[width=\textwidth]{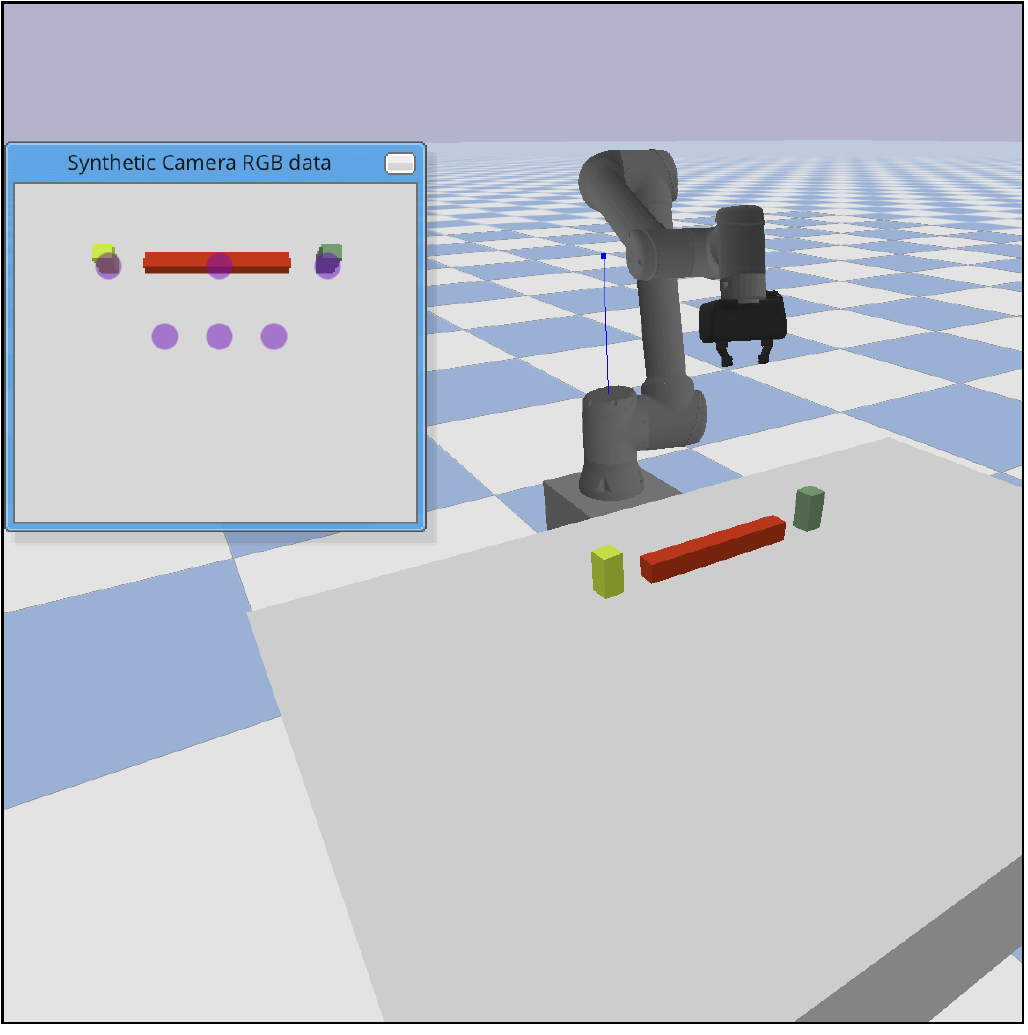}
        \caption{The experiment setup.}
        \label{subfig:setup}
    \end{subfigure}
    \hspace{6em}
    \begin{subfigure}[b]{0.3\textwidth}
        \centering
        \includegraphics[width=\textwidth]{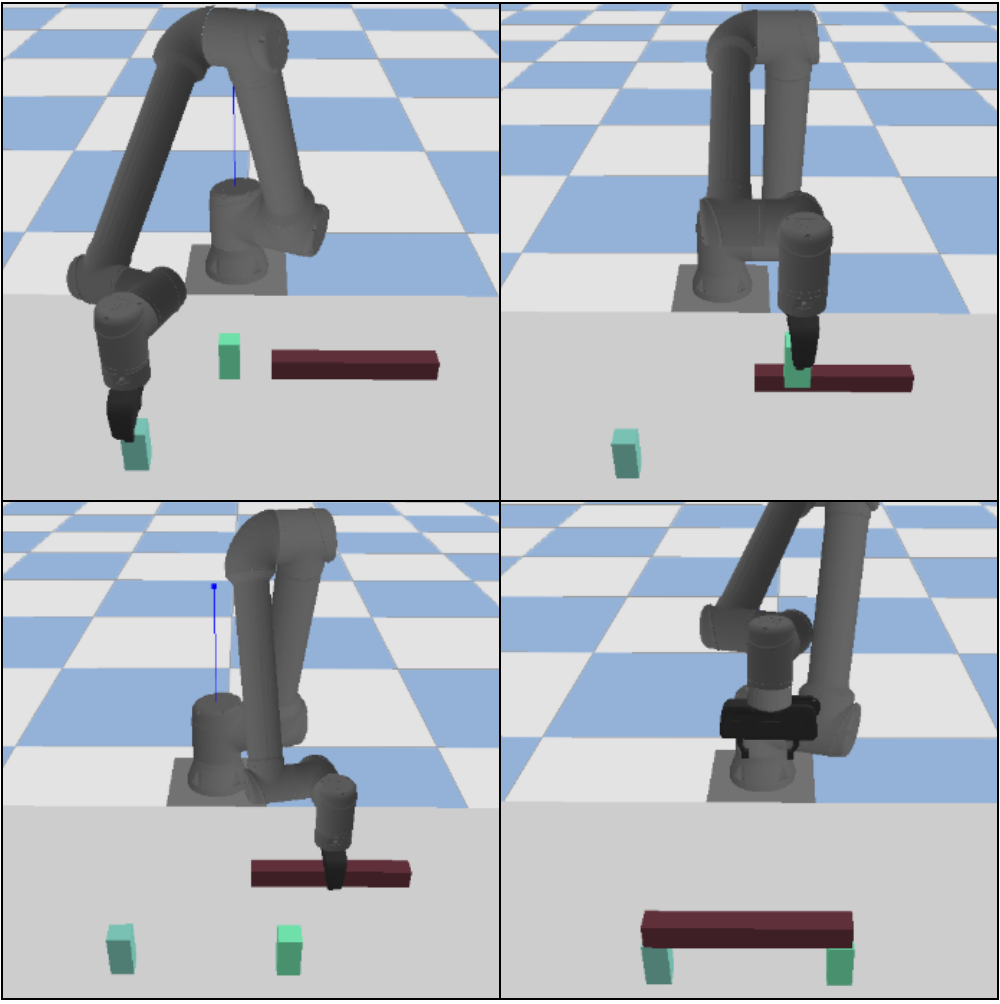}
        \caption{Example exploration.}
        \label{subfig:exploration}
    \end{subfigure}
    \caption{The experiment setup. Six possible pick and release locations are shown in purple in the synthetic camera image in \hyperref[subfig:setup]{(a)}. An example exploration with three objects is shown in \hyperref[subfig:setup]{(b)}.}
    \label{fig:exp_setup}
\end{figure*}

The proposed architecture is shown in Fig.~\ref{fig:model}. This architecture predicts the effects of an executed action at a given state. The architecture consists of an encoder $f(x)$ with a binarized bottleneck layer, a decoder $g(x)$, and an attentive module $a(x)$. We assume that we have an image segmentation module for cropping and tracking objects through several timesteps. While this is a strong assumption, state-of-the-art segmentation methods in computer vision can segment and track objects in complex environments \cite{locatello2020object,elsayed2022savi++}.

In the first part, the high-dimensional continuous state $x$ is first partitioned into segments $(x_1, x_2, \dots, x_k)$ using the segmentation module. As we expect to obtain relational symbols that potentially encode information related to several objects, the relative positions of the objects are important and need to be preserved while processing image segments. In order not to lose the location information, here we concatenate the location of each pixel in the cropped image as two additional channels.
Then, each segmented image ($x_i$) is given as input to the encoder. The encoder consists of several convolutional layers for processing high-dimensional visual input. For binarization at the last layer of the encoder, we use the Gumbel-sigmoid function \cite{maddison2016concrete,jang2016categorical} for gradient backpropagation. In the end, the encoder outputs a binary vector $z_i$ for each segment $x_i$. These binary vectors ($z_i$) are expected to encode symbols that can encode one or more objects and that are effective in predicting the effects of actions and, therefore, multi-step planning.

Each action is represented with a one-hot binary vector. Actions are assumed to be high-level parameterized motion primitives. The action vector $a$ is concatenated with each $z_i$ separately. Concatenated vectors $((z_1, a), (z_2, a), \dots, (z_k, a))$ are given as input to the attentive module. The attentive module consists of several self-attentive layers \cite{vaswani2017attention} which allows the generated symbols $(z_1, z_2, \dots, z_k)$ to interact with each other to generate $(h_1, h_2, \dots, h_k)$ that should hold relational information for accurate effect prediction. Note that we assumed the existence of action primitives in this work. These primitives can be learned with different motion primitive methods \cite{paraschos2013probabilistic,schaal2006dynamic,seker2019conditional} and transferred from previous stages of development as we previously showed \cite{ugur2015staged}.

The self-attention operation allows each symbol to attend to other symbols to form a new representation. Since the whole model is trained in an end-to-end fashion, symbols are formed in such a way that they not only define the characteristics of the cropped object but also contain information about relations with other objects. For example, if a long stick is picked and released to a different location, and if there is a small cube on top of the stick, then the position of the cube changes as well. To accurately predict such relational effects, the model needs to encode information regarding relations between objects. Self-attention seamlessly combines this information using query-key-based attention operation proposed in \cite{vaswani2017attention}.

As the last step, the decoder function predicts the generated effect for each segment. In our case, we represent the effect eff$(x_i)$ as the position displacement of the $i$th object after the action. In our experiments, we directly retrieve the position displacement from the simulator. However, one can use a more generic effect such as the change in pixel values as in \cite{ahmetoglu2020deepsym}. The learned symbols heavily depend on the effect representation, and we treat effect representation as a separate problem. This work, instead, focuses on creating an architecture that can learn symbols for a varying number of inputs and predict action effects that would require processing relational information between objects.

\section{EXPERIMENTS}
\subsection{Experiment Setup}
The experiment setup is shown in Fig.~\ref{fig:exp_setup}. This is a tabletop environment where a UR10 robot arm picks up and releases objects at six pre-defined locations shown in Fig.~\ref{subfig:setup}. Given six pick-up and six release locations, the total number of high-level actions is 36. An example sequence of action executions is shown in Fig.~\ref{subfig:exploration}. The environment is initialized with one to three objects initially located at the row closer to the robot body (see the top three purple dots in Fig.~\ref{subfig:setup}). The robot perceives its environment with a depth camera located on top of the table. The camera takes $256 \times 256$ pixels depth image. We assume that the robot has a segmentation module that can crop and track the movement of objects. This can be achieved with state-of-the-art slot-based models \cite{locatello2020object,elsayed2022savi++}. In this work, we use the segments provided by the simulator. An example crop is shown in Fig.~\ref{fig:model} (i). All crops are $64 \times 64$ pixels. Finally, in order to preserve object locations, the $x$ and $y$ locations of each pixel in the cropped image are concatenated as two additional channels.

The robot collects the interaction data set as follows. The environment is initiated with $k\in \{1,2,3\}$ objects where $k$ is set randomly. The initial depth image $x_i$ of the environment together with its segmentation $s_i$ is recorded before taking any action. Then, a random action $a_i$ is executed and position displacements of $k$ objects $e_i = ([\Delta x_1, \Delta y_1, \Delta z_1], \dots, [\Delta x_k, \Delta y_k, \Delta z_k])_i$ are recorded as the generated effects. Here, $x$ and $y$ represent the pixel coordinates of the object's center of mass, and $z$ represents the depth value obtained from the depth camera for the corresponding object. Note that the length of $e_i$ depends on the number of objects. 
In total, 12,000 $(x_i, s_i, a_i, e_i)$ tuples are recorded as the interaction data set. We use 10,000 samples for training, 1,000 samples for validation, and the remaining 1,000 samples for testing.

We train the architecture in Fig.~\ref{fig:model} to predict the generated effects $e_i$ of an action $a_i$. The encoder $f(x)$ consists of four convolutional layers with 64, 128, 256, and 512 filters. Convolutional layers are similar to that of DCGAN \cite{radford2015unsupervised} with a kernel size of four, a stride of two, and a padding of one. After convolutions, we take an average across height and width dimensions, and project this fixed-size vector into an 8-dimensional vector. Lastly, we binarize the activations using Gumbel-sigmoid function for backpropagation \cite{jang2016categorical,maddison2016concrete}. The self-attention module $a(x)$ is a transformer with four transformer encoder layers \cite{vaswani2017attention}. We use the default settings in PyTorch for transformer layers \cite{paszke2017automatic}. The decoder $g(x)$ is a multi-layer perceptron (MLP) with three hidden layers, each containing 256 units. We use batch-normalization \cite{ioffe2015batch} in the encoder and the decoder to increase the convergence speed. We train all modules in an end-to-end fashion with Adam optimizer \cite{kingma2014adam}.

\begin{figure}[btp]
    \centering
    \includegraphics[width=0.9\columnwidth]{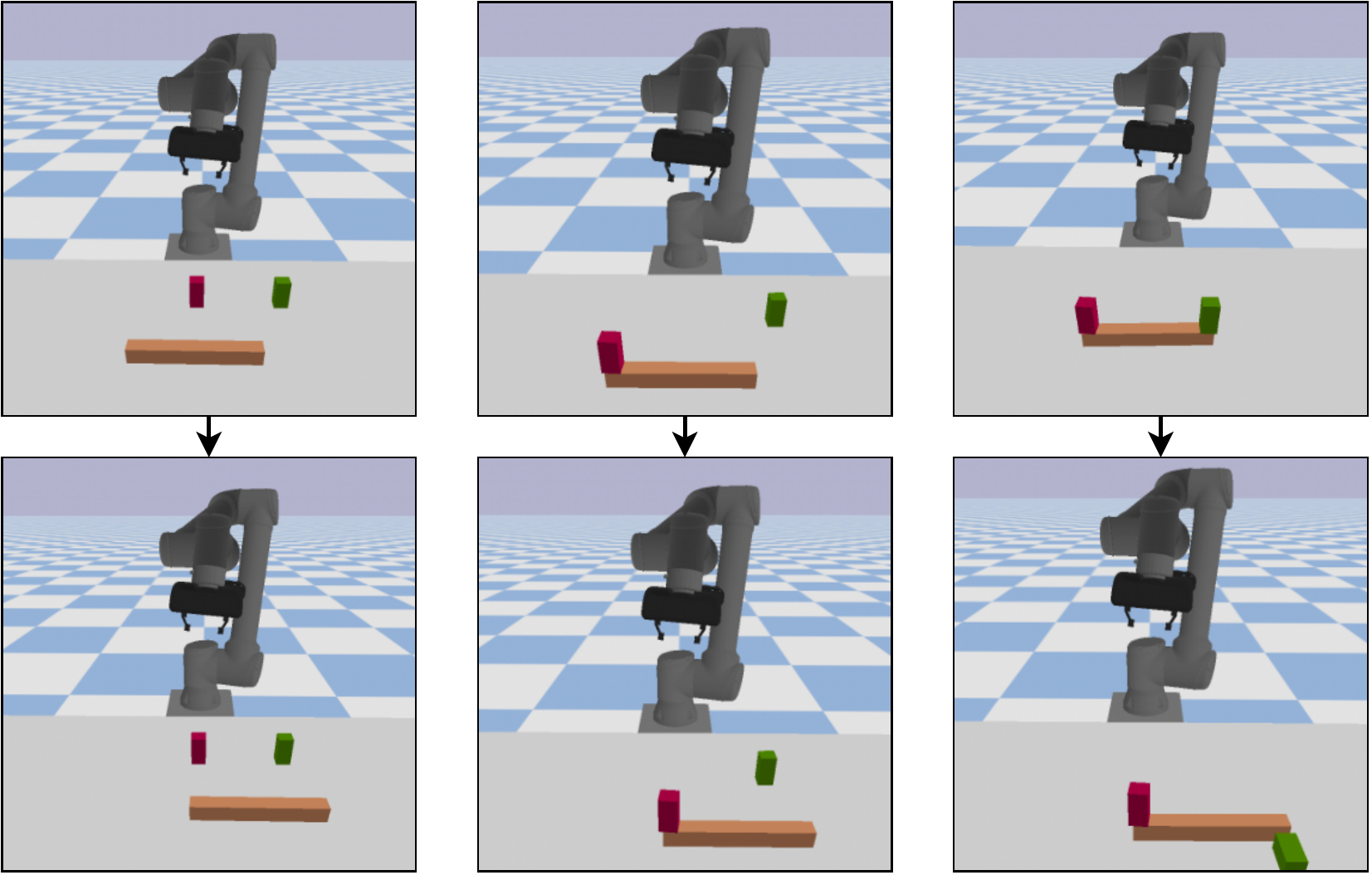}
    \caption{Effect prediction for different states.}
    \label{fig:effect_example}
\end{figure}

\subsection{Effect Prediction}

Our system predicts the continuous effect, i.e., continuous pixel position and depth changes of objects, from learned discrete symbolic activations. It is neither our aim nor possible to minimize the prediction error to zero, yet we need to make sure that our system can make discover symbols that can make fairly good effect predictions. Therefore, we analyzed the prediction error. After training, the mean effect prediction errors on the test set are 8.6mm, 15mm, and 6mm for $x$, $y$, and $z$ dimensions, respectively. For comparison, the average changes of the corresponding dimensions of objects in the effect set are 50mm, 68mm, and 10mm. As a result, we can conclude that our system discovered symbolic representations that are effective in predicting effects of actions on single or multiple objects.

\begin{table}[hbtp]
    \centering
    \caption{Effect predictions for example cases in Fig.~\ref{fig:effect_example}. Units are in millimeters.}
    \label{tab:effect_results}
    \begin{tabular}{lc|c|c}
        \hline
        & Green Cube & Orange Stick & Red Cube \\
        \hline
        \multicolumn{4}{c}{Predictions} \\
        \hline
        Case 1 & (3, 6, -2) & (-2, 192, -2) & (4, -9, -1) \\
        Case 2 & (3, 8, 0) & (-17, 196, 1) & (-10, 180, -5) \\
        Case 3 & (-3, 210, -45) & (4, 159, 8) & (-30, 131, -20)\\
        \hline
        \multicolumn{4}{c}{Ground truth} \\
        \hline
        Case 1 & (0, 0, 0) & (0, 180, 0) & (0, 0, 0) \\
        Case 2 & (0, 0, 0) & (0, 180, 0) & (-1, 176, 0) \\
        Case 3 & (102, 177, -75) & (3, 180, 0) & (2, 176, 0)\\
        \hline
    \end{tabular}

\end{table}

Next, we investigate whether our system discovered symbols that automatically include information from action-relevant objects in multi-object settings, and model interaction dynamics of these objects. For this, we created a scenario where the action applied to an object is the same, but the other objects in the environment are arranged in different ways such that different effects are expected to be obtained and are expected to be correctly predicted by our system. The action is to pick up a long stick from one position and release it to another position, and two other objects are placed in different configurations, as shown in Fig.~\ref{fig:effect_example}. In Fig.~\ref{fig:effect_example}, three different example interactions experienced by the robot were shown in three columns. The initial snapshot of each interaction is shown in the upper row, and the final snapshot after action execution is shown in the bottom row. In all three cases, the robot picks up the orange stick, moves it to one position right, and releases it (see the bottom-right purple position in Fig.~\ref{subfig:setup}). Due to different initial configurations of the red and green blocks, different effects are (expected to be) observed.

The effects predicted by our system along with ground-truth effect values are provided in Table~\ref{tab:effect_results} for each interaction (case). The results show us that our system was able to model relational information between objects and therefore made correct predictions. For example, in Case 1, the system correctly predicted that only the position of the stick changes. In Case 2, the position of the red cube also changes along with the stick. And finally, in Case 3, the positions of both cubes change along with the stick (in the same direction). This shows that our system learned symbols that enable it to make high-level reasoning that involves multiple objects, such as ``objects on top of another object will move together with the object below''. We conclude that self-attention in transformer layers indeed helps the binary symbols to interact with each other to predict the correct effect. In the previous work, DeepSym \cite{ahmetoglu2020deepsym}, the modeling of such interactions was only possible at the input level by manually concatenating the necessary object crops with a pre-defined number of objects. Here, due to the self-attention layers, the model discovers symbols that automatically use the binary activations of the related objects and disregards the activations of unrelated objects. This is a clear advantage when compared with the DeepSym architecture.

\begin{figure*}[t]
    \centering
    \includegraphics[width=0.9\textwidth]{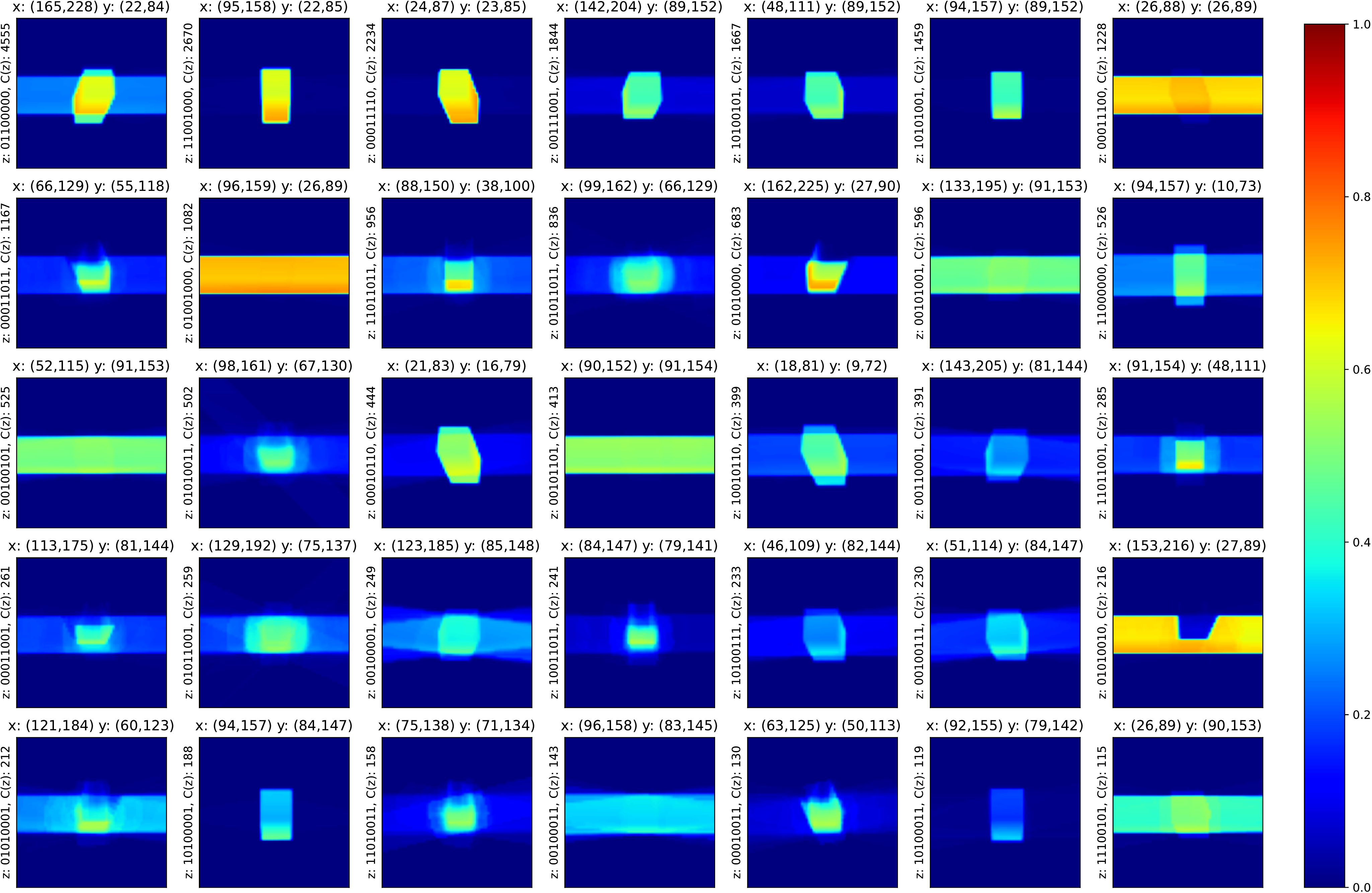}
    \caption{The average depth images for each symbol sorted by their activation count. Only the symbols that are activated more than 100 times out of 28,607 samples (the number of objects in 10,000 samples) are shown in the figure. The symbol and its activation count are reported on the left, and the average x-axis and y-axis pixel locations are reported on the top of each subfigure. The color bar shows the normalized depth value.}
    \label{fig:symbols}
\end{figure*}

\begin{figure}[thbp]
    \centering
    \includegraphics[width=0.8\columnwidth]{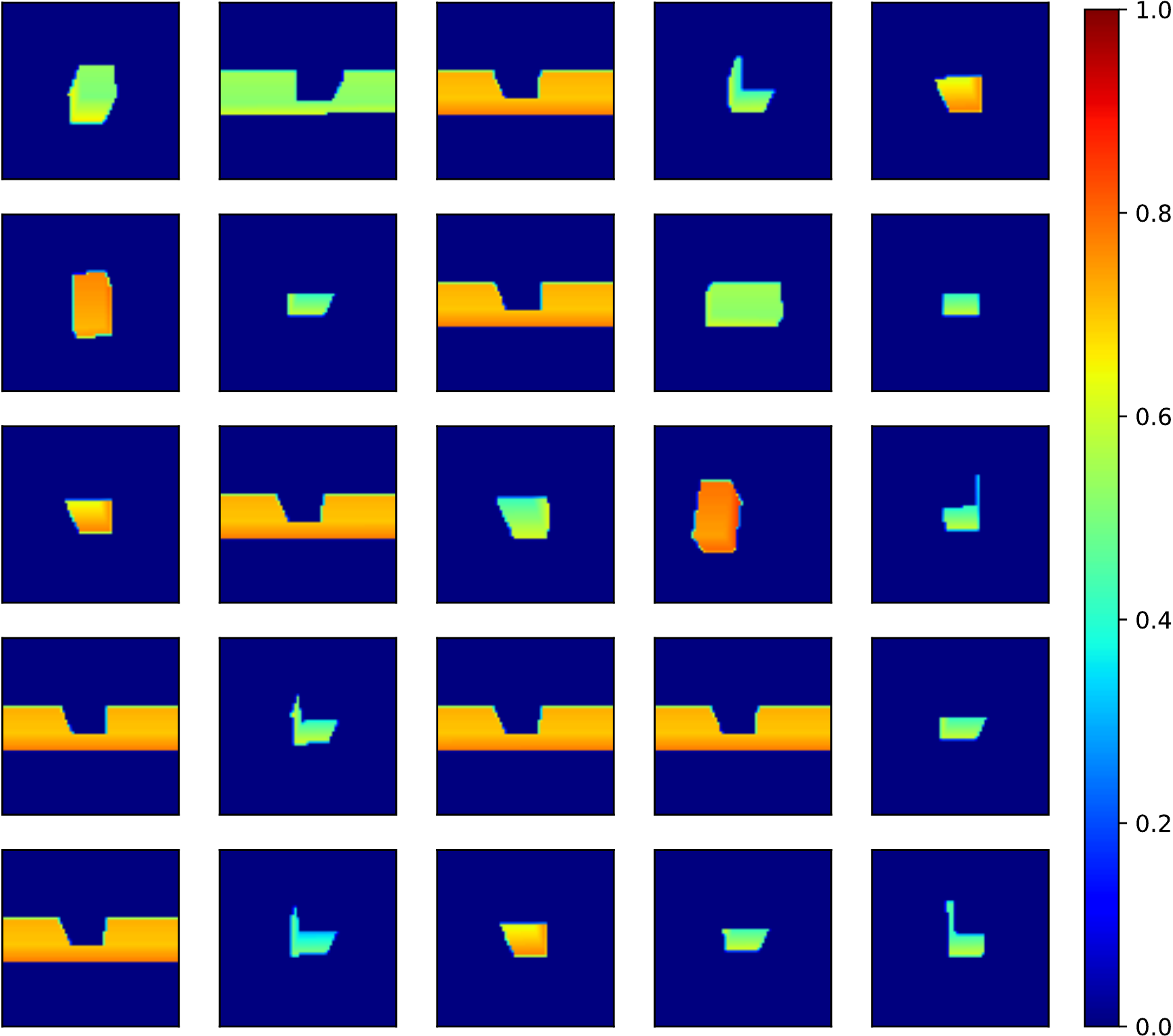}
    \caption{Example images that activate the symbol `00011011'. The common feature in these examples is that the object is occluded by another object. Notice that this symbol is activated for both sticks and cubes. Since the robot cannot grasp the occluded object and therefore cannot change its position. This symbol can rather be interpreted as `not graspable'.}
    \label{fig:example_symbol}
\end{figure}

\subsection{Learned symbols}
The number of binary units in the post-encode bottleneck layer is set to 8 units; therefore, there can be at most 256 unique symbols. As we analyzed the activation frequency of these symbols, we observed that the most frequently activated 35 symbols covers 95\% of the training set. The prototypical values, i.e., the average states from samples that activate these 35 specific object symbols, are shown in Fig.~\ref{fig:symbols}. As shown, these object symbols encode the location of the object, the depth of the object, and the occlusion information.

While a number of symbols are activated only for sticks or only for cubes, others are activated for both cubes and sticks. For example, the symbol in the second row and the first column is activated for both cubes and sticks (see samples in Fig.~\ref{fig:example_symbol}). This makes sense since the generated effect is the same in both cases; the robot cannot grasp the occluded object, and therefore, cannot change its position. We can interpret this as a `not graspable' symbol. However, note that there is not a specific symbol for `not graspable stick' or `not graspable cube' since it does not bring any additional advantage in terms of effect prediction accuracy; the model can already predict the generated effect (which is 0mm, 0mm, 0mm on average) without differentiating these two objects. We can conclude that our system learned a minimal set of symbols that is required to make predictions and reasoning, and these symbols were not only determined by the available objects in the environment but also by the action capabilities of the robot.

\subsection{Planning with Discovered Symbols}

\begin{figure}[tbp]
    \centering
    \includegraphics[width=0.8\columnwidth]{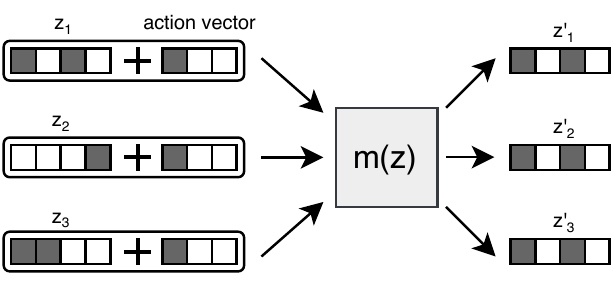}
    \caption{Symbolic forward module. Given the symbolic representation of the current state, this module directly predicts the symbolic representation of the next state.}
    \label{fig:symbol_forward}
\end{figure}

In this section, we investigate the suitability of symbolic representations in making multi-step plans. Although our system can predict the effects on objects given the tabletop image and the action, here, we aim to acquire a full symbolic reasoning capability by learning and exploiting a transition model that predicts the next symbolic state given the current symbolic state and the action. For this, we train a separate network $m(z)$ that predicts the next symbolic state given the current symbolic state and the action (see Fig.~\ref{fig:symbol_forward}) using the discovered symbolic representations that were presented in the previous sections. This network consists of two dense layers followed by a single self-attention layer and another two dense layers. We train the network with binary cross entropy loss for each dimension. 

\begin{figure}
    \centering
    \includegraphics[width=\columnwidth]{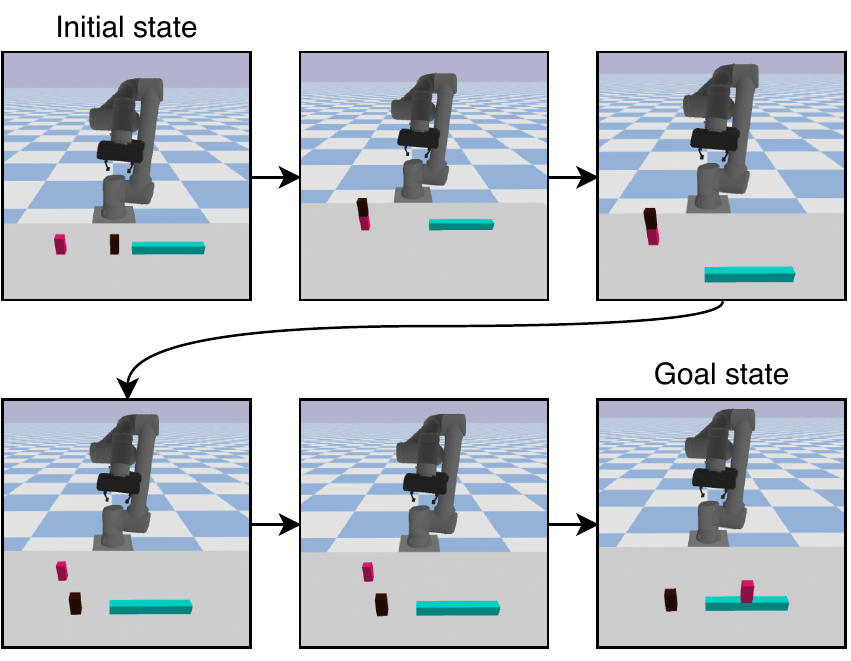}
    \caption{An example plan for an arbitrary goal state. In the fourth action, the robot tries to pick and release from the top-right position to the bottom-left position, which does not interact with any object.}
    \label{fig:plan1}
\end{figure}

Such a network, after training, allows us to search for a goal symbolic state using any tree search algorithm \cite{russell2010artificial}. Fig.~\ref{fig:plan1} shows a plan generated and executed in order to achieve a composite structure given as a goal. 
As shown, the learned symbolic transition model could be used to generate feasible plans and, when executed, could achieve the desired goal, indicating that multi-step predictions in the generated long horizon plan were correct.
This example shows that the discovered symbols can be used for training a forward symbol prediction model that allows achieving goals via tree search.

\section{CONCLUSION}
We proposed and implemented a predictive encoder-decoder network that utilized a binary bottleneck layer and, importantly, a self-attention mechanism in order to discover symbols relevant for interacting with the environment of the robot that hosts a varying number of objects. Through experiments with a simulated manipulator robot, we showed that the robot acquired reasoning capabilities to encode interaction dynamics of a varying number of multiple objects in different configurations using the discovered symbols. For example, when queried, the robot could reason that (possible multiple numbers of) objects that are on top of another object would move together if the object below is picked-up, and the objects around would not move. We showed that these reasoning capabilities were acquired by learning a minimal set of symbols that are optimized for effect prediction in the ecological niche of the robot. We also showed that the discovered symbols can be used for planning by training a higher-level neural network that makes pure symbolic reasoning. In order to acquire optimal plan generation capability, we plan to incorporate effective and efficient plan generation requirement in the symbol discovery loop as suggested by \cite{silver2022inventing}. We also plan to transfer the learned symbolic reasoning capabilities to the real robot via domain randomization techniques.





\section*{ACKNOWLEDGMENT}
This research was supported by TUBITAK (The Scientific and Technological Research
Council of Turkey) ARDEB 1001 program (project number: 120E274) and by the BAGEP Award of the Science Academy. The numerical calculations reported in this paper were partially performed at TUBITAK ULAKBIM, High Performance and Grid Computing Center (TRUBA resources). Additional support is provided by the International Joint Research Promotion Program, Osaka University under the project ``Developmentally and biologically realistic modeling of perspective invariant action understanding” and the Japan Society for the Promotion of Science, Grant-in-Aid for Scientific Research - project number 22H03670.

\bibliographystyle{IEEEtran}
\bibliography{IEEEabrv,ref}

\end{document}